\documentclass[Afour,sageh,times]{sagej}

\usepackage{moreverb,url}

\usepackage[colorlinks,bookmarksopen,bookmarksnumbered,citecolor=red,urlcolor=red]{hyperref}

\newcommand\BibTeX{{\rmfamily B\kern-.05em \textsc{i\kern-.025em b}\kern-.08em
T\kern-.1667em\lower.7ex\hbox{E}\kern-.125emX}}

\def\volumeyear{2022}

\begin{document}

\runninghead{ALTO: A Large-Scale Dataset for UAV Visual Place Recognition and Localization}

\title{ALTO: A Large-Scale Dataset for UAV Visual Place Recognition and Localization}


\author{Ivan Cisneros\affilnum{*}, Peng Yin\affilnum{*}, Ji Zhang\affilnum{*}, Howie Choset\affilnum{*}, Sebastian Scherer\affilnum{*}}

\affiliation{\affilnum{*}Robotics Institute, Carnegie Mellon University, USA.\\
}

\corrauth{Peng Yin, Robotics Institute, Carnegie Mellon University, Pittsburgh, PA 15213, USA.}
\email{pyin2@andrew.cmu.edu}

\begin{abstract}
We present the ALTO dataset, a vision-focused dataset for the development and benchmarking of Visual Place Recognition and Localization methods for Unmanned Aerial Vehicles. The dataset is composed of two long (approximately $150$km and $260$km) trajectories flown by a helicopter over Ohio and Pennsylvania, and it includes high precision GPS-INS ground truth location data, high precision accelerometer readings, laser altimeter readings, and RGB downward facing camera imagery. In addition, we provide reference imagery over the flight paths, which makes this dataset suitable for VPR benchmarking and other tasks common in Localization, such as image registration and visual odometry. To the author's knowledge, this is the largest real-world aerial-vehicle dataset of this kind. Our dataset is available at \href{https://github.com/MetaSLAM/ALTO}{https://github.com/MetaSLAM/ALTO}.
\end{abstract}


\maketitle

\section{Introduction}
While Unmanned Aerial Vehicles (UAVs) primarily rely on GPS-assisted navigation, there is a need for active and passive sensor -assisted navigation in order to ensure reliability of the robot. 
In order to be resilient to GPS dropout or spoofing, it is important to have a fallback navigation system that is reliable, robust, and onboard. Vision-focused navigation techniques are of interest to us because of the many advantages cameras have over active sensors such as Lidar: cameras are low-cost, low-weight, and low-power, which are all characteristics that are well-received on UAV platforms, which are known to be constrained in all of these areas. 

The ability of mobile robots to recognize previously visited or mapped areas is essential to achieving reliable autonomous systems. Place recognition for localization is a promising method for achieving this ability. However, current approaches are negatively affected by differences in viewpoint and environmental conditions that affect visual appearance (e.g. illumination, season, time of day), and so these methods struggle to provide continuous localization in environments that change over time. These issues are compounded when trying to localize in large-scale (several-kilometer length) maps, where the effects of repeated terrain (visual aliasing) result in greater localization uncertainty and false positive data associations. While there exist many Visual Place Recognition (VPR) datasets that are large-scale~\cite{Torii2018-be} and/or multi-domain~\cite{Torii2013-fq}, few, are tailored to the UAV domain, and fewer still can be used for navigation and SLAM related tasks.

The ALTO (\textbf{A}erial-view \textbf{L}arge-scale \textbf{T}errain-\textbf{O}riented) dataset aims at filling this gap and pushing forward the development of UAV specific VPR and vision-focused navigation techniques.

This dataset was collected using a custom sensor payload mounted at the nose of a Bell helicopter. The dataset consists of two large-scale flights ($150$km and $260$km) over a wide variety of different terrain (forest, urban, rural, rivers, lakes, plains) in the Ohio and Pennsylvania regions. The camera is nadir-facing and not gimbaled and provides an unobstructed view of the terrain that is being flown over. Both flights are captured in the same day in the summer.

In addition to the large amount of downward-facing aerial imagery captured on this flight, we provide preprocessed geo-referenced orthoimagery that can be used as a database for the development of VPR methods.


\begin{figure}[t]
	\centering
    \includegraphics[width=\linewidth]{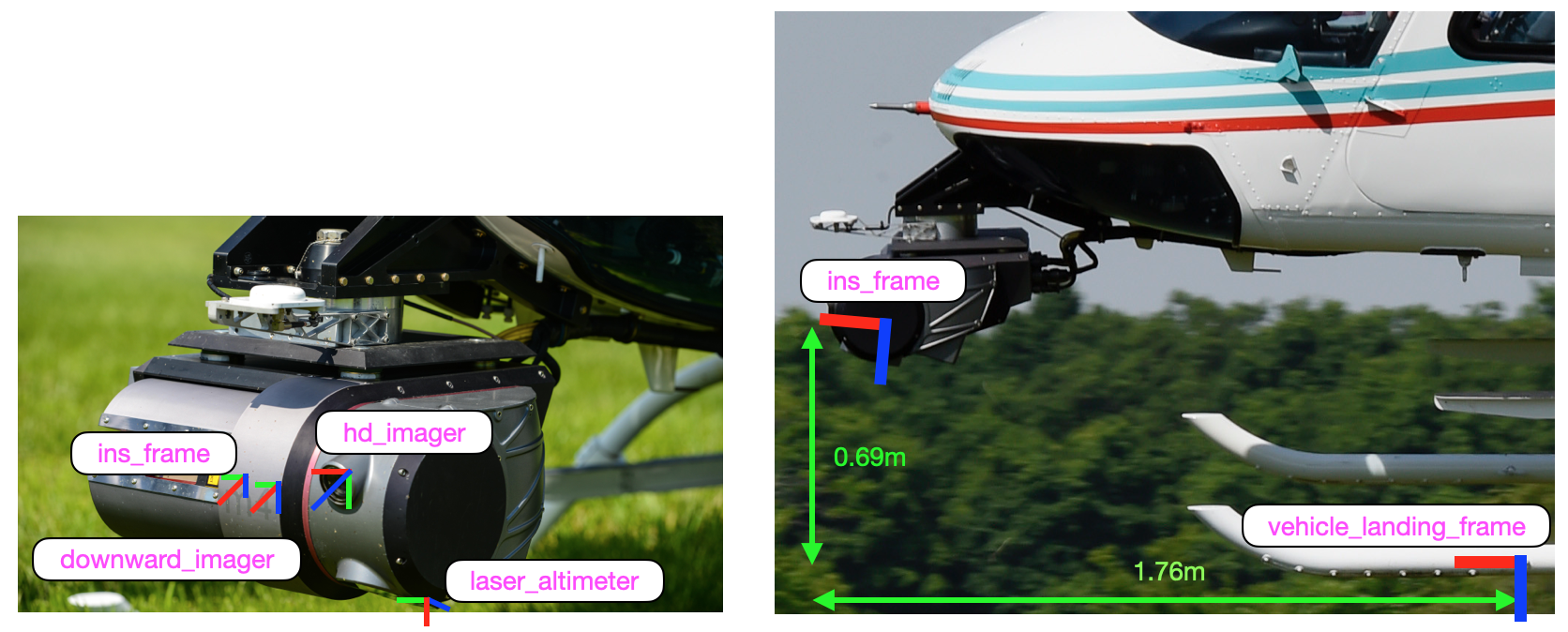}
	\caption{\textbf{The custom sensor payload and its position on the data capture vehicle.}}
	\label{fig:platform}
\end{figure}

\newpage
\section{The Platform}
\subsection{Sensors}
The sensor payload setup is illustrated in Figure \ref{fig:platform}. The sensors are:
\begin{enumerate}
  \item Downward imager: The camera is an IDS Imaging UI-5250RE-C, which has a 1.92 Megapixel CMOS color sensor (1600 x 1200 resolution)~\cite{ids:ui5250}. The camera has a global shutter, and the exposure and white balance were both set to auto. It is fitted with a Kowa LM3NC1M 3.5mm lens ~\cite{kowa}.  We collected the imagery in Bayered format at 20 Hz.
  \item GPS-INS system: The GPS receiver is a NovAtel SPAN OEMV, paired with a LITEF LCI-1 IMU.  The combined system has single point 1.5m Horizontal Position Accuracy (RMS) ~\cite{novatel}. This data was captured at 200 Hz.
  \item Laser Altimeter: The downward-facing laser altimeter is a Loke LMC-J-0310-1~\cite{loke}. It is configured to have a max range of 700m, with a resolution of 1mm. We capture this data at 20 Hz.
\end{enumerate}

\subsection{Coordinate Frames}
The origin of our data coordinate system is the standard ECEF frame. The ‘ins\_frame’ is the root frame that we use for the payload; this frame is oriented with the z-axis down (towards the Earth), x-axis in the forward direction (towards the nose of the helicopter).  We provide the GPS-INS pose (position and quaternion) and the IMU (velocities and acceleration) data as transformations of the ins\_frame frame with respect to the ECEF frame. The ECEF coordinates and orientations of the sensor frames on the payload can be derived from their relationships to the ins\_frame. We provide a .npy which contains these relative transforms. In addition to the GPS-INS data in the ‘ins\_frame’, we also provide a separate file with logs of an ‘ltf’ frame at each timestamp; this frame represents the orientation of the NED tangent plane, but it is translated to be located at the same position as the ‘ins\_frame’.  This can be useful, for example, for reorienting an image and removing all roll, pitch, and yaw relative to the NED tangent plane.  

An important note is that the ‘ltf’ or NED orientation has the x-axis pointing northwards, whereas the camera frame ‘downward\_imager’ has the x-axis pointing towards the nose of the helicopter; a properly north-aligned image will, thus, need an additional 90 degree clockwise rotation around the ‘ltf’ z-axis.

\section{The Dataset} 
\subsection{Overview}
\begin{figure}[t]
	\centering
    \includegraphics[width=\linewidth]{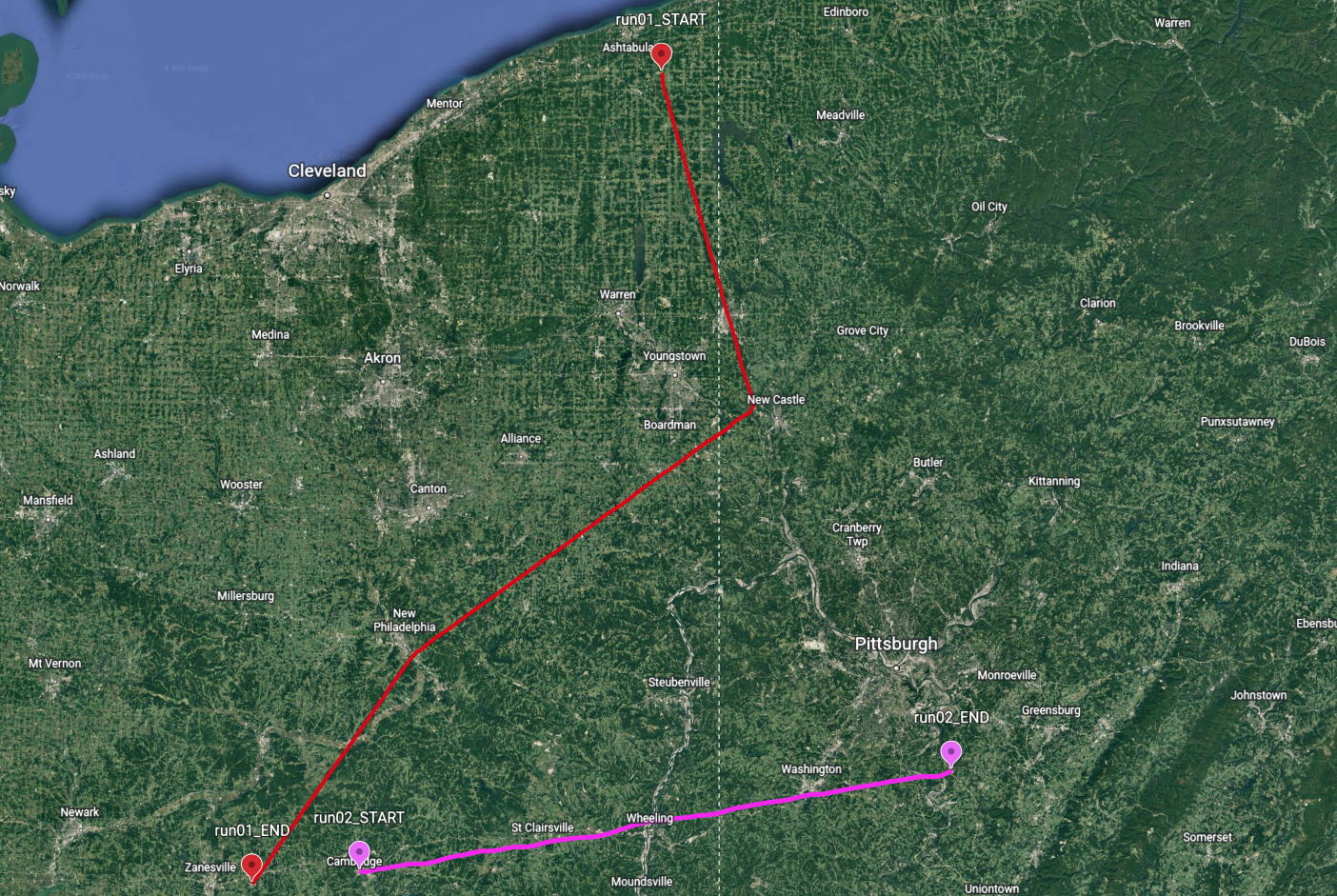}
	\caption{\textbf{The dataset trajectories.}
    Here we show the entire length of the two trajectories collected in this dataset. ``run\_01'' (red) runs primarily from North to South.  ``run\_02'' (magenta) runs primarily from West to East. Both are located in UTM zone 17N.}
	\label{fig:run01_run02_wide_view}
\end{figure}

The dataset is composed of two flights, designated ``run\_01'' and ``run\_02''. The trajectories are visualized in Figure \ref{fig:run01_run02_wide_view}. From start to end, trajectory ``run\_01'' is $260$km long, and ``run\_02'' is $150$km long.  In the data, we do not include the ascent and descent phases of the flights. Trajectory ``run\_01'' starts at Northeast Ohio Regional Airport (lat: $41.7778$, long: $-80.6876$), and ends at  Zanesville VOR-DME (lat: $39.9471$, long: $-81.8898$).  Trajectory ``run\_02'' starts at Cambridge Muni Airport (lat: $39.9806$, long: $-81.572$) and ends at  Rostraver Airport (lat: $40.2093$, long: $-79.8348$) right outside of Pittsburgh.  The two flights were captured in the same day in the summer, thus lighting and foliage should be similar between the two sets of imagery.

The reference orthoimagery was obtained from the National 
Agricultural Imagery Program (NAIP)  of the U.S. Department of Agriculture~\cite{usgs}. We preprocess and stitch together the GeoTIFFs that cover the regions traversed in our two trajectories. 

The current dataset is available at the link given in the Abstract, and is limited to only a few kilometers for the General Place Recognition Competition 2022.  The full dataset is forthcoming.

\bibliographystyle{unsrtnat}
\bibliography{references}





\end{document}